\documentclass[a4paper,10pt,twoside,twocolumn,final]{JICS_LaTexGuidelines} 

\usepackage[fleqn]{amsmath} 
\usepackage[pdftex]{graphicx} 
\usepackage{amssymb,amsfonts} 
\usepackage{algorithmic}
\usepackage{textcomp}
\usepackage{xcolor}
\usepackage{booktabs}
\usepackage{multirow}
\usepackage{multicol}
\usepackage{url}
\usepackage{balance} 
\usepackage{soul}
\usepackage{xspace}

\usepackage{geometry} 
\newlength{\bibitemsep}
\newlength{\bibparskip}
\let\oldthebibliography\thebibliography
\renewcommand\thebibliography[1]{%
  \oldthebibliography{#1}%
  \setlength{\parskip}{\bibitemsep}%
  \setlength{\itemsep}{\bibparskip}%
}

\begin{document}
\title{Evolutionary Algorithms in Approximate Computing: A Survey} 

\author{Lukas Sekanina\textsuperscript1
\\\vspace{1.2em}\small\textsuperscript1Brno University of Technology, Faculty of Information Technology, Bozetechova 2, 612 66 Brno, Czech Republic\\ e-mail: sekanina@fit.vutbr.cz
} 
%
\markboth{Submitted to Journal of Integrated Circuits and Systems}%
{SEKANINA L.: Evolutionary Algorithms in Approximate Computing: A Survey}%
\renewcommand*\footnoterule{}%
\footnotetext{DOI}%
\maketitle
\begin{abstract}
In recent years, many design automation methods have been developed to routinely create approximate implementations of circuits and programs that show excellent trade-offs between the quality of output and required resources. This paper deals with evolutionary approximation as one of the popular approximation methods. The paper provides the first survey of evolutionary algorithm (EA)-based approaches applied in the context of approximate computing. The survey reveals that EAs are primarily applied as multi-objective optimizers. 
We propose to divide these approaches into two main classes: (i) parameter optimization in which the EA optimizes a vector of system parameters, and (ii) synthesis and optimization in which EA is responsible for determining the architecture and parameters of the resulting system. The evolutionary approximation has been applied at all levels of design abstraction and in many different applications. The neural architecture search enabling the automated hardware-aware design of approximate deep neural networks was identified as a newly emerging topic in this area.
\end{abstract}  
\begin{indexterms}
evolutionary algorithm; approximate computing; digital circuit; neural network; optimization.
\end{indexterms} 

\section{Introduction}

As Dennard’s scaling is no longer valid for the newest technologies, it is still more and more difficult to increase the performance of computer systems under a given power budget~\cite{Hipeac:visio:2021}. Moreover, challenging technological issues (such as higher process variability) typical for the smallest technology nodes lead to reliability problems that must be properly handled during manufacturing, testing, verification, and design. One of the most prominent approaches developed to address these issues is \emph{approximate computing}~\cite{Mittal:2016,ACsurvey:ACM:2020}. It exploits the fact that many applications are intrinsically error-resilient. This resilience can be utilized to simplify their implementation and thus reduce power consumption, area, or delay for an acceptable loss in the quality of output. Current pervasive applications such as machine learning, image, video and speech processing, data mining, and robotics, fortunately, belong to this class of error-resilient applications. 

Many design automation methods have been developed to create approximate implementations for this important class of applications
~\cite{Mittal:2016,ACsurvey:ACM:2020,Scarabottolo:pieee:2020}. State-of-the-art results were quite often obtained by \emph{evolutionary algorithms} (EAs)~\cite{Sekanina:bookch:19}. The reason for the good applicability of EAs for purposes of approximate computing is that EAs are excellent in stochastic multi-objective design and optimization and in handling various constraints. The approximation problem can be formulated as a {\em multi-objective optimization} problem in which the accuracy, performance, and resources are conflicting design objectives. EA can be seeded by existing solution(s) whose functionality is degraded in a controlled way in the course of evolution. The result of the evolution process is a set of solutions that exhibit useful trade-offs among the key observed objectives. 

In the context of approximate computing, EAs have been applied at different levels of design abstraction, for optimization as well as synthesis purposes, and in numerous applications. The goal of this paper is to provide the first systematic survey of the utilization of EAs in approximate computing. We identified over 60 papers in this area that we divided into two main classes concerning the use of EA: (i) \emph{parameter optimization} in which the EA optimizes a vector of system parameters, and (ii) \emph{synthesis and optimization} in which the EA is responsible for determining the architecture and parameters of the resulting system. As evolutionary algorithms are frequently used for purposes of design, optimization, and approximation of deep neural networks (DNNs), we handle this topic separately from the previous two in the third part of this survey, which deals with the \emph{evolutionary approximation in neural networks}. After reading this paper, the reader should understand why, when, and how EAs are currently used in the context of approximate computing.  

The rest of the paper is organized as follows. Sections~\ref{sec:axc} and \ref{sec:eas} provide an introduction to approximate computing and evolutionary algorithms, respectively. In Section~\ref{sec:classif}, we introduce our approach to the classification of the evolutionary approaches in approximate computing. Section~\ref{sec:po} is then devoted to the applications in which an EA optimizes a vector of parameters of a system undergoing the approximation. Section~\ref{sec:so} deals with the evolutionary design of approximate implementations by genetic programming. The evolutionary approximation of neural networks is discussed in Section~\ref{sec:nnaprox} Conclusions are given in Section~\ref{sec:conclusions}

\section{Approximate Computing}
\label{sec:axc}

According to Mittal~\cite{Mittal:2016} ``\emph{Approximate computing exploits the gap between the level of accuracy required by the applications/users and that provided by the computing system, for achieving diverse optimizations}". The error resilience originates in applications with (i) analog inputs (such as image processing, sensor data processing, and voice recognition that operate on noisy real-world data), (ii) analog output (which is intended for human perception and can inherently tolerate errors imperceptible to users),  (iii) no unique answer, and (iv) iterative processing of large amounts of data where the iterations can be stopped earlier or approximated using a heuristics~\cite{Hadi:CACM:2015}.

Approximate computing has been applied at different levels of the computing stack. Surveys~\cite{Mittal:2016,ACsurvey:ACM:2020} provide detailed introduction to relevant approximation techniques. At the lowest level, \emph{voltage over-scaling} enables to reduce power consumption while some timing errors are introduced into the circuit. One of the most popular and general-purpose approximation techniques is \emph{functional approximation} which replaces the exact circuit or program by a simplified implementation. The functional approximation is often used in approximate implementations of arithmetic circuits as well as accelerators. In the basic case, it takes a form of a suitable bit width reduction or pruning of less critical parts of the circuit. However, sophisticated strategies based on logic re-synthesis have also been proposed~\cite{Scarabottolo:pieee:2020}. 
Approximations can be injected to the memory subsystem by approximating the cache functionality, memory operations, or elementary memory cells. At the highest level of abstraction, approximations can be introduced at the level of the instruction set architecture, schedulers, compilers, or to the source code through techniques such as loop perforation, task skipping, relaxed synchronization, and memorization.

Approximate implementations are usually created manually by a domain expert. However, a recent trend is to develop fully automated approximation methods capable of (i) identifying suitable components that should undergo the approximation process with the highest priority, (ii) generating candidate approximate implementations, and (iii) evaluating them concerning various criteria. These methods are constructed using iterative heuristics~\cite{Scarabottolo:pieee:2020}.

The quality of approximate implementations is typically expressed using one or several error metrics, where the most commonly used ones are the error rate (ER), the mean absolute error (MAE), and the worst-case error (WCE):
\begin{align} 
\mathrm{ER} &= \frac{\sum_{\forall x \in B: O_\mathrm{approx}(x) \neq O_\mathrm{orig}(x)} 1}{n} \\[0.5em] 
\mathrm{MAE} &= \frac{\sum_{\forall x \in B} \left|O_\mathrm{approx}(x) - O_\mathrm{orig}(x)\right|}{n} \\[0.5em] 
\mathrm{WCE}  &= \max_{\forall x \in B} \left|O_\mathrm{approx}(x) - O_\mathrm{orig}(x)\right| 
\label{eq:error6}
\end{align}
where the output of the approximate implementation and original (exact) implementation is $O_\mathrm{approx}$ and $O_\mathrm{orig}$, $B$ contains all possible input vectors, and $n = |B|$.
Apart from these general-purpose metrics, the error can also be evaluated at the application level using application-specific metrics such as the Peak Signal to Noise Ratio (PSNR) for image processing. The error is obtained either by simulation (typically only a subset of $B$ is considered to reduce a considerable computation overhead), error probability modeling~\cite{Mazahir:2017aa} or exact formal analysis~\cite{vasicek:access2019}. For example, in the case of 16-bit approximate multipliers generated in the course of approximation, the relative error between the exact WCE and the WCE estimated using $10^8$ randomly generate vectors (out of all possible $2^{32}$ vectors) can go far beyond 10\%. This inaccuracy of simulation has motivated the development of exact formal error analysis methods that are surveyed, e.g., in ~\cite{vasicek:access2019}. 

\section{Evolutionary Algorithms}
\label{sec:eas}
Evolutionary algorithms are metaheuristic search algorithms that employ a population of candidate solutions and bio-inspired operators such as mutation, crossover, and selection to search effectively in the space of candidate solutions~\cite{EibenS15}. Contrasted to the gradient-based optimization methods, nothing is supposed about the objective function except its evaluability.

A generic single-objective EA utilizing \emph{fitness function} $f$, which assigns a fitness score to every candidate solution, works as follows. After initializing and evaluating the first population, the algorithm repeats the following steps until a termination condition is not satisfied: (i) selection of parents for a new population; (ii) creating offspring from the parents by means of crossover, mutation, and other relevant operators; (iii) creating the new population by considering parents as well as offspring; (iv) evaluation of the new population. The candidate solution with the highest fitness score is the result of EA (if the objective is to maximize the fitness).

Various branches of EAs primarily differ in problem encoding, genetic operators, and fitness function construction. These topics are briefly discussed in the following paragraphs. At the same time, we have to omit some advanced topics (such as co-evolutionary algorithms, parallel EAs, and various hybrid methods) and related algorithms (ant colony optimization (ACO), particle swarm optimization (PSO), etc.) because of limited space.

\subsection{Types of EAs} 

The \emph{genetic algorithm} (GA) tries to optimize a vector of parameters. This vector is called genotype, its length is usually constant, and the values it contains are either of the same data types (e.g., Boolean values, integers, or floats) or mixed data type. GA uses fitness proportional or tournament selection (for details, see~\cite{EibenS15}) to determine the parents of the new population. Mutation randomly modifies each item of the parental vector (the so-called gene) with a very low probability. Crossover swaps some parts of two parents and creates two offspring. The new population is composed of either  offspring individuals (a generational GA) or both the offspring and parents (a steady-state GA). 


\emph{Genetic programming} (GP) has been invented for automated program synthesis~\cite{poli08:fieldguide}. Candidate programs are represented by syntax trees. GP defines a set of terminals (the inputs to the program and constants) and a set of functions that are used as building blocks of candidate programs. The search algorithm is almost identical to the generic EA. In addition to crossover and mutation, more complex genetic operators are often employed, e.g., subprogram definition and reuse~\cite{poli08:fieldguide}. In order to obtain the fitness score, a candidate program $\alpha$ is executed on a training data set $D_{tr}$ containing $n$ pairs $(x_i, y_i)$, where $y_i$ is the expected output value for the input $x_i$. The objective of GP is to minimize the overall error, which is typically expressed in the fitness function as 
\begin{equation}
f(\alpha, D_{tr}) = \sum_{i=1}^{n} E(y_i, \alpha(x_i)),
\end{equation}
where $E$ is a suitable error metric. The result of GP (i.e., the best performing program at the end of evolution) is evaluated on test data to validate its performance. GP is often presented as a solver for \emph{symbolic regression problems}.

\emph{Cartesian Genetic Programming} (CGP) was established for the evolutionary circuit design~\cite{Miller:cgp:2020}. A candidate solution is represented using a grid of $n_c \times n_r$ nodes, each of them implementing one of $n_a$-input functions from the function set. A candidate circuit has $n_i$ primary inputs and $n_o$ primary outputs. 
The genotype consists of $(n_a+1)n_{c}n_{r} + n_o$ integers that define (i) function codes of the nodes, (ii) the connections among the nodes and with primary inputs, and (iii) the connection of the primary outputs. The feedback connections are not allowed in the basic version of CGP. The genotype thus specifies a directed acyclic graph (some nodes of  the grid can remain unused). CGP applies a point mutation as its basic genetic operator and a simple $(1+\lambda)$ search strategy, in which $\lambda$ offspring are generated from a single parent. 
The main advantage of CGP is that subcircuits can easily be reused, and a designer can constrain the maximum size of the circuit by a suitable setting of the number of columns. Furthermore, CGP can evolve solutions at various levels of abstraction, including the transistor, gate, register-transfer (RT) as well as program level, by choosing suitable types of nodes~\cite{Miller:cgp:2020}.

\emph{Linear Genetic Programming} (LGP) focuses on the automated design of instruction-level programs for processors with general-purpose registers~\cite{poli08:fieldguide}. Some registers hold the program inputs, and the program output is expected in predefined register(s). A candidate solution is a machine-level program consisting of instructions, each of them in the form $(OP, R_a, R_b, R_c)$, where OP is the operation code and $R_a$, $R_b$, and $R_c$ are registers used by this instruction. Candidate programs are encoded as strings of integers. The mutation is quite standard as it alters a randomly selected integer to any permitted value. The crossover is usually applied at the level of instructions, not individual genes.

\subsection{Multi-Objective Optimization}

The approximation problem can be seen as a \emph{multi-objective optimization problem}, i.e. an optimization problem that involves \emph{multiple objective functions} $g_1(a), g_2(a), \dots, g_k(a)$, where $g_i: \mathbb{A} \to \mathbb{R}$, $k$ is the number of objectives, and $a \in \mathbb{A}$ is a candidate solution from the set of solutions $\mathbb{A}$. In the multi-objective optimization, there does not typically exist one solution that minimizes all objective functions simultaneously because the design objectives are conflicting. Hence, rather than one (optimal) solution, the optimization results in a set of solutions, i.e. the solutions that cannot be improved in any of the objectives without degrading at least one of the other objectives. Formally, a solution $a$ is said to \emph{(Pareto) dominate} another solution $b$, $a, b \in \mathbb{A} $, if:\\
$g_{i}(a)\leq g_{i}(b)$ for all $i\in \left\{{1,2,\dots ,k}\right\}$\\ 
and\\
$g_{j}(a) < g_{j}(b)$ for at least one index $ j\in \left\{{1,2,\dots ,k}\right\},$\\
and all $g_i$ have to be minimized. 
A solution $a^{*} \in \mathbb{A}$  is called a \emph{non-dominated solution} if there does not exist another solution that dominates it. The set of non-dominated solutions is called the \emph{Pareto front}. We say that non-dominated solutions are \emph{Pareto optimal solutions} if all possible candidate solutions are considered during the optimization, and there are no provably better non-dominated solutions in the search space. In practice, we are almost always faced with a situation in which the EA produces suboptimal solutions, i.e., the resulting Pareto front contains the best non-dominated solutions obtained during the evolution, but not truly Pareto-optimal solutions. A recent survey of evolutionary multi-objective optimization methods is in~\cite{moea:2020}. 

A common approach adopted in the design of approximate implementations is either (i) transforming the multi-objective problem to a single-objective one (using suitable constraints, prioritization or aggregation methods) and solving it with a common single-objective method or (ii) employing a truly multi-objective approach. 

When using \emph{constraints}, only one of the objective functions is optimized while the remaining ones are transformed to constraints $g_i(a) \leq c_i$, where $c_i$ are suitable constants. A \emph{penalty function} is then introduced to punish any solution violating the constraints. However, this method does not guarantee that all hard constraints are satisfied because solutions with a high penalty can be promoted to the next generation if there are no better solutions. The \emph{prioritization} means that the most important objective is optimized first, and when a suitable solution is obtained, the second most important objective is optimized but ensuring that the first one is not worsened. This is repeated for all the objective functions according to their priority. Finally, the \emph{aggregation methods} introduce a suitable aggregation function (such as the weighted sum, weighted exponential sum, or weighted product) for the objective functions and optimize the composition. This approach suffers from several problems. First, it is not easy to determine suitable weights. Second, the linear weighted sum only works for problems with convex Pareto fronts, i.e., solutions on non-convex segments are unreachable. Third, similar to the constrained optimization, the method has to be executed several times with different weight settings to approximate the Pareto optimal front.

Truly multi-objective optimization methods iteratively build the Pareto front in the course of optimization by comparing candidate solutions using the non-dominance relation, promoting good solutions, and trying to cover the expected Pareto front. In comparison with other multi-objective optimization methods, EAs gained a significant attention in this task. One of the most popular methods is NSGA-II~\cite{deb2002}. It is based on sorting individuals in a population according to the dominance relation into multiple fronts. The first front contains all non-dominated solutions. Each subsequent front is constructed by removing all the preceding fronts from the population and finding a new Pareto front in the remaining individuals. The solutions within the individual fronts are then sorted according to the crowding distance metric. This metric helps to preserve the diversity of the population along the fronts. Best individuals from these fronts then serve as parents for the new population. NSGA-II thus always produces a set of solutions (i.e., a Pareto front) when terminated.

\section{Proposed Classification}
\label{sec:classif}

In this section, we present \emph{four observations} that we utilized to introduce the proposed classification of EAs in the context of approximate computing.

Employing the EA is natural concerning its goal in the approximation task. Minor modifications introduced in the course of evolution into existing solutions (note that EA is typically seeded with an exact solution) and the principle of the survival of the fittest naturally lead to discovering such solutions which show very good trade-offs between the error and other objectives. 
Our {\bf first observation} is that all problems solved by EAs in the context of approximate computing are multi-objective optimization problems. Hence, it makes no sense to distinguish between single- and multi-objective approaches.

The {\bf second observation} is related to the problem encoding. The inspected papers can be divided into two main classes. The {\bf Parameter Optimization} class deals with problems in which the user specifies a set of parameters that have to be optimized. These parameters are then stored in the genotype, and a GA is typically employed to optimize their values. A typical optimization problem within this class is a component selection problem in which the GA's task is to select the most suitable approximate component (from a library of components) for each of $n$ predefined positions in the circuit or program implementation. On the other hand, the {\bf Synthesis and Optimization} class covers such problems in which approximate programs or circuits are synthesized and optimized. This task is performed by a suitable type of GP. This is a more challenging problem than the parameter optimization because the genotype encodes not only types of components but also their connection, and the search space is thus much larger. 

One of the fitness functions typically evaluates the error of candidate solutions. Other fitness functions are devoted to performance (delay, latency) and resources (area, power). The {\bf third observation} reflects the fact that the fitness score can be either precisely evaluated (which is usually computationally demanding) or estimated (which is less expensive). 
See Section~\ref{sec:axc} for a quick overview of the approaches adopted for the error calculation. 
Regarding the other objectives: in the course of evolution, it is not tractable to precisely evaluate all circuit parameters (in particular, power consumption) using professional design tools. Hence, basic circuit characteristics are \emph{estimated} in the fitness function, and only resulting circuits are fully evaluated using the standard process at the end of evolution. 

Our {\bf fourth observation} is that the evolutionary approximation has been applied at all levels of the circuit design abstraction, i.e., at the level of abstract models (such as binary decision diagrams (BDDs)), gates, look-up tables (LUT), register transfer (RTL), and behavioral description, as well as at the level of software and compilers. Both the Parameter Optimization and the Synthesis and Optimization methods have been utilized on all levels of design abstraction. 

Table~\ref{tab:class:poso} shows that we classified the papers into two main classes: {\bf Parameter Optimization} and {\bf Synthesis and Parameter Optimization}. Within these classes we further use the following sub-classes: the {\bf EA} used -- GA, GP, CGP, LGP, PSO; {\bf Level} of abstraction -- gate, LUT, RTL, SW, and BDD; {\bf Error calculation} method -- simulation, statistical method, BDD analysis, SAT solving, combinatorial analysis; {\bf Platform} -- processor (CPU), microcontroller (MCU), field programmable gate array (FPGA), and application-specific integrated circuit (ASIC). Table~\ref{tab:class:poso} also briefly characterizes the {\bf Application domain} and {\bf Benchmark problems} for each paper.

Because of limited space, we do not provide any systematic comparison between EAs and other approximation methods. Only in some cases, we highlight some interesting improvements obtained by EAs with respect to other methods. More details are available in referenced papers. 

\begin{table*}[t]
    \centering
    \caption{Selected papers in which the EA is used for purposes of approximate computing. Specific abbreviations: Sim -- simulation; Statistical -- statistical error analysis; Sim$+$Comb -- Simulation and combinatorial analysis.}
    \label{tab:class:poso}
    \resizebox{\textwidth}{!}{
    \begin{tabular}{lllllllll} 
\hline
{\bf Year}	&	{\bf Ref.}	&	{\bf EA}	&	{\bf Level}	&	{\bf EA's target}	&	{\bf Error Calc.}	&	{\bf Application domain}	&	{\bf Benchmark problems}	&	{\bf Platform}	\\
\hline 
\multicolumn{9}{c}{\bf Parameter Optimization}\\
\hline
2011	& \cite{Ansel:2011} &	GA	&	SW	&	config. parameters	&	Sim	&	SW auto-tuning	&	5 programs	&	CPU	\\
2014	& \cite{Park2014ExpAXAF} &	GA	&	SW	&	program variables	&	Sim	&	SW approximation	&	9 programs	&	CPU	\\
2015	& \cite{demara:15} &	GA	&	gate	&	adder selection	&	Sim	&	arithm. circuit	&	adder	&	ASIC	\\
2016	& \cite{Grater:DATE16} &	GA	&	SW	&	porgram variables	&	Sim	&	OpenCL kernels	&	6 programs	&	FPGA	\\
2016	& \cite{Vaverka:CompLib:16} &	GA	&	RTL	&	component selection	&	Statistical	&	high level synthesis	&	DCT, reduce sum	&	ASIC	\\
2017	& \cite{ApproxCompressSens:2017} &	GA	&	RTL	&	multiplier selection	&	Sim	&	image compression	&	4 images	&	ASIC	\\
2018	& \cite{ApproxMultNair:18} &	GA	&	gate	&	adder selection	&	Sim	&	multiplier design	&	edge detection, k-means clustering	&	ASIC	\\
2018	& \cite{Albandes:2018} &	GA	&	gate	&	gate selection	&	Sim	&	TMR	&	5 benchmark circuits	&	ASIC	\\
2019	& \cite{Kadiyala:19} &	GA	&	RTL	&	component selection	&	Sim	&	FIR filter bank	&	hearing aid	&	ASIC	\\
2020	& \cite{CANDARW:20} &	GA	&	RTL	&	config. parameters	&	Sim	&	display rendering pipeline	&	image color improvement	&	FPGA	\\
2021	& \cite{WorkloadAware:DATE2021} &	GA	&	RTL	&	component selection	&	Sim	&	component selection	&	Sobel, Prewitt, and Laplacian filter	&	ASIC	\\
2021	& \cite{BarbareschiBM21} &	GA	&	RTL	&	bit widths	&	Sim	&	bit width assignment	&	spam detector	&	FPGA	\\
2021	& \cite{Barone:Access:21} &	GA	&	SW	&	parameters, components	&	Sim	&	SW, cross-layer	&	K-Means, Taylor, JPEG, Sobel, CNN	&	FPGA	\\
\hline
\multicolumn{9}{c}{\bf Synthesis and Optimization}\\
\hline
2013	& \cite{Petrlik:ddecs13} &	CGP	&	RTL	&	circuit synthesis	&	Sim	&	multiple constant multiplier	&	5 multipliers	&	none	\\
2013	& \cite{sekanina:axc:ices13} &	CGP	&	gate	&	circuit synthesis	&	Sim	&	logic synthesis	&	cm152, sym9, t481, 9-majority, 3b and 4b adders	&	ASIC	\\
2015	& \cite{vasicek:sekanina:tec} &	CGP	&	gate/RTL	&	circuit synthesis	&	Sim	&	logic synthesis	&	2b, 3b, 4b multiplier, 9-median, 25-median	&	ASIC	\\
2015	& \cite{Mrazek:gecco:GI:2015} &	CGP	&	SW	&	program synthesis	&	Sim	&	SW code	&	9-median, 25-median	&	MCU	\\
2016	& \cite{vasicek:sekanina:genp16} &	CGP	&	gate	&	circuit synthesis	&	BDD	&	logic synthesis	&	16 circuits	&	ASIC	\\
2016	& \cite{Vas:Mra:Sek:ICES16} &	CGP	&	gate	&	circuit synthesis	&	Sim	&	logic synthesis	&	8b adder, 8b multiplier, median, Sobel, Gaussian filters	&	FPGA	\\
2016	& \cite{TREL:16} &	CGP	&	gate	&	circuit synthesis	&	Sim	&	TMR	&	9 circuits	&	ASIC	\\
2016	& \cite{Vasicek:fpl:16} &	CGP	&	gate/LUT	&	circuit synthesis	&	BDD	&	logic synthesis	&	24 circuits	&	FPGA	\\
2017	& \cite{BDDapprox:gecco:17} &	GA	&	BDD	&	variable permutation	&	BDD	&	logic synthesis	&	20 circuits 	&	none	\\
2017	& \cite{axcdct:17} &	CGP	&	RTL	&	circuit synthesis	&	Sim	&	DCT	&	3 images	&	ASIC	\\
2017	& \cite{Vasicek:Mrazek:genp17} &	CGP	&	SW	&	program synthesis	&	Sim$+$Comb	&	median	&	sensor data filter, image filters	&	MCU	\\
2017	& \cite{Vasicek:date:hevc} &	CGP	&	gate	&	circuit synthesis	&	BDD	&	adder	&	DCT in HEVC	&	ASIC	\\
2017	& \cite{Mrazek:date17:evoapprox8b} &	CGP	&	gate	&	circuit synthesis	&	Sim	&	arithm. circuits	&	8b adder, 8b multiplier	&	ASIC	\\
2017	& \cite{Ceska:iccad17} &	CGP	&	gate	&	circuit synthesis	&	SAT	&	arithm. circuits	&	up to 32b multipliers, up to 128b adders	&	ASIC	\\
2017	& \cite{Sekanina:radioeng17} &	CGP	&	RTL	&	circuit synthesis	&	Sim	&	image filter	&	30 images	&	ASIC	\\
2018	& \cite{Grochol:ahs18} &	LGP	&	SW	&	program synthesis	&	Sim	&	hash function	&	network flow hashing	&	FPGA	\\
2018	& \cite{GPforFEx:TC18} &	GP	&	SW	&	program synthesis	&	Sim	&	features extraction programs	&	EEG-seizure, ECG-cardiac-arrhythmia	detection &	ASIC	\\
2018	& \cite{Wiglasz:ICES18} &	CGP	&	RTL	&	program synthesis	&	Sim	&	sqrt, arctan	&	feature extraction	&	CPU	\\
2018	& \cite{Mrazek:iet:18} &	CGP	&	cell	&	circuit synthesis	&	Sim	&	arithm. circuits	&	12b multiplier	&	ASIC	\\
2018	& \cite{Mrazek:AHS18} &	CGP	&	gate	&	circuit synthesis	&	Sim	&	quality configurable circuit	&	Gaussian noise filter	&	ASIC	\\
2019	& \cite{Kemcha:Nedjah:19} &	PSO	&	RTL	&	circuit synthesis	&	Sim	&	sequential circuit design	&	divider	&	ASIC	\\
2019	& \cite{Vasicek:date2019} &	CGP	&	gate	&	circuit synthesis	&	Sim	&	arithm. circuit	&	Gaussian noise filter, CNN	&	ASIC	\\
2020	& \cite{ReducingArea:2020} &	GP	&	SW	&	program synthesis	&	Sim	&	features extraction programs	&	EEG-seizure, ECG-cardiac-arrhythmia detection	&	ASIC	\\
2020	& \cite{Souza:MOCGPApprox:2020} &	CGP	&	gate	&	circuit synthesis	&	Sim	&	logic synthesis	&	15 circuits	&	ASIC	\\
2020	& \cite{Ceska:ASC:2020} &	CGP	&	gate	&	circuit synthesis	&	SAT	&	arithm. circuits	&	32b multiplier, 32b MAC, 24b divider	&	ASIC	\\
2021	& \cite{Vasicek:DDECS:21} &	CGP	&	LUT	&	circuit synthesis	&	Sim	&	arithm. circuits	&	up to 64b adders and 8b, 16b multipliers 	&	FPGA	\\
\hline
 \end{tabular}
    }
\end{table*}

\section{Parameter Optimization}
\label{sec:po}

The parameter optimization problem consists of optimizing $n$ parameters that the user must carefully choose from all the relevant parameters associated with a given system under approximation. These parameters typically determine the bit width of components, the version of approximate implementation of a component, or other configuration options. The genotype then contains $n$ values that are optimized by EA with the aim of finding the best trade-offs between the error and other objectives.

\subsection{Hardware Level}

The following papers demonstrate that the parameter optimization can be introduced at different levels of abstraction.

Several approximate transistor-level implementations of a one-bit full adder are considered in~\cite{demara:15}. A parallel genetic algorithm is then employed to select the most suitable ones to serve as building blocks of multi-bit adders. 

Following the same idea, from a set of different approximate adder blocks, a GA is used in~\cite{ApproxMultNair:18} to select the most relevant ones to build a heterogeneous multiplier. At the same time, the MAE is minimized for the overall design. 

In \cite{ApproxCompressSens:2017}, NSGA-II performs this selection from a library of 11 approximate implementations to minimize the power-area product and error in an image compression application. Compared to the baseline architecture that uses regular multipliers in the 65~nm CMOS technology, 43\% area, and 54\% power savings were obtained with a minimal PSNR degradation. 

In~\cite{WorkloadAware:DATE2021}, GA tries to select the most suitable versions of approximate components (adders and multipliers) for approximate implementations of Sobel, Prewitt, and Laplacian filter. The goal is to minimize energy consumption while meeting the quality constraints. 

Barbareschi et al.~\cite{BarbareschiBM21} employs NSGA-II to find the best trade-offs between the classification accuracy and hardware cost for a set of classifiers (based on decision trees) that are implemented in FPGA. The genotype specifies the number of bits needed for each comparison conducted in the first layer of the classifier. 

A multi-objective evolutionary algorithm is employed to solve the so-called binding problem of high-level synthesis (HLS). The task is to find an optimal assignment of approximate components to nodes of the data flow graph describing a complex digital circuit~\cite{Vaverka:CompLib:16}. The genotype contains $n$ integers specifying a particular component version assuming that the data flow graph has $n$ nodes for which the biding must be determined.  The error is estimated using a statistical error analysis based on error propagation models in arithmetic circuits. The method is evaluated using the reduce (sum) and discrete cosine transform (DCT) circuits. As the library of approximate circuits (a predecessor of EvoApprox8b~\cite{Mrazek:date17:evoapprox8b}) was too large (hundreds of approximate components), two methods enabling to reduce its volume (and so the search space) while still providing suitable trade-offs for the HLS are proposed and compared. 

And finally, in this category, we present two more sophisticated approaches. In \cite{Kadiyala:19}, the parameter optimization is combined with constrained pruning. The objective is to approximate a digital signal processing block of a hearing aid consisting of $m$ FIR filter banks. The method utilizes a library of pruned circuit topologies and approximate multipliers indexed according to the level or degree of pruning.  
The approximation problem is formulated as assigning a particular inexact level to each filter bank (i.e., all components of the bank will have the same degree of approximation). The genome contains $m$ integers that define the level of approximation for each bank. The goal of NSGA-II is to minimize power consumption and maximize the quality of signal processing. The resulting inexact FIR filter bank is 1.92$\times$ or 2.56$\times$ more efficient in terms of power consumed while producing 10\% or 20\% less intelligible speech, respectively, compared with a hearing-aid utilizing exact filters.

A display rendering pipeline (including circuits for tone mapping, color space conversion, electro-optical transfer function compensation, etc.) is constructed to process input images in monitors. This pipeline can be characterized using several parameters whose values can be optimized to maximize the image color quality while minimizing the area and power consumed on an FPGA. These parameters are encoded into a genotype containing thus the number of fraction bits, types of approximate adders, etc. and optimized by NSGA-II~\cite{CANDARW:20}. 

\subsection{Software Level} 

In the context of software-level approximation, the parameter optimization can take various forms.

Ansel et al.~\cite{Ansel:2011} propose a new kind of language extension and an accuracy-aware compiler to create a code supporting various degrees of accuracy for the PetaBricks programming language and compiler. The compiler performs empirical auto-tuning based on a GA to build an optimized configuration for each accuracy level required by a user. GA maintains a population of candidate algorithm configurations which is expanded using a set of mutators. The mutator functions are different for each program and are generated fully automatically with information from static analysis. Mutators can modify configuration variables or decision trees that decide which algorithm to use for each choice site, accuracy, and input size. To evaluate candidate programs, the number of tests is dynamically specified with the aim of shortening the evaluation time and, at the same time, obtaining a statistically relevant fitness score. The method was evaluated using six benchmarks that are representative of commonly used algorithms that leverage variable accuracy. 

ExpAX~\cite{Park2014ExpAXAF} introduces annotations to the source code that enable to express error expectations. For example, the expectation\\
\verb, magnitude(v) > c using E with rate < d, 
allows to bound both the error rate and the error magnitude of variable \verb,v,: it states that the rate at which the error incurred on variable \verb,v, exceeds normalized magnitude \verb,c, is bounded by \verb,d, concerning the error metrics \verb,E,. A static safety analysis is performed that uses the high-level expectations to infer a safe-to-approximate set of program operations automatically. This step requires a finding of possible safe-to-approximate variables. Unsafe-to-approximate variables are variables violating memory safety or functional correctness. A GA is used to determine a subset of safe-to-approximate operations that minimizes the error and energy requirements. The genotype is a bit vector indicating for each variable whether it is or is not suitable for approximation. Significant energy savings (up to 35\%) with the considerable reduction in programmer effort (3$\times$ to 113$\times$ fewer annotations w.r.t EnerJ~\cite{EnerJ:SampsonDFGCG11}) were reported while providing formal safety and statistical quality-of-result guarantees.

GRATER~\cite{Grater:DATE16} performs a sensitivity analysis to find safe-to-approximate variables in OpenCL kernel. The genome specifies the precision (i.e., data type) of these variables. The objective is to find an approximate kernel that minimizes the resource utilization on FPGA while meeting the target quality. 

Barone et al.~\cite{Barone:Access:21} propose E-IDEA, an automatic application-driven approximation tool targeting different implementations (hardware and software). E-IDEA uses Clang-Chimera tool to analyze the Abstract Syntax Tree (AST) of the C/C++ source code of the application. Through the so-called mutators, approximations can be introduced at the level of the source code. The set of mutators includes loop-perforation mutators,  precision-scaling mutators for the floating-point arithmetic, a precision scaling mutator for the integer arithmetic, and a mutator supporting approximate arithmetic operator models of circuits being part of the EvoApproxLib library~\cite{Mrazek:date17:evoapprox8b}. An evolutionary approximation method based on NSGA-II tries to find the best approximation version of a given C/C++ code, according to user-defined optimization objectives. A candidate solution is represented as a vector of integers; each of them corresponds to one parameter that can be mutated. The positions in the source code at which a mutation can be applied are specified by a set of matching rules. The method is evaluated using five applications (K-Means, Taylor, JPEG, Sobel, and CNN), for which it shows good trade-offs between the error and hardware cost.
	
\section{Synthesis of Approximate Implementations}
\label{sec:so}

This section deals with evolutionary algorithms that optimize not only some circuit or program parameters, but also simplify, optimize, or even redesign circuit topology. This task is predominantly solved by genetic programming, and by CGP in particular. We will start our survey with gate-level designs and then continue up to higher levels of abstraction. CGP usually begins with an exact circuit implementation and tries to find the best trade-offs between the error and other properties of the approximate implementations. In the multi-objective setup, CGP is either combined with NSGA-II (or another truly multi-objective EA) or some of the objectives are considered as constraints (e.g., the error and delay) while CGP minimizes the remaining one (e.g., the power consumption). The CGP-based approximation is often utilized for arithmetic circuits, but the method is identical if a general logic synthesis (and optimization) problem is considered. Contrasted to the conventional paradigm, which employs different methods for the synthesis of arithmetic circuits and general logic, this unified methodology is seen as a strong advantage of CGP. 


\subsection{Gate Level Approximations}

The first work on evolutionary gate-level approximation is from 2013~\cite{sekanina:axc:ices13}. CGP is used to minimize the error while resources (the number of gates) available to CGP are constrained. Interesting trade-offs are evolved for four single-output circuits and 2-bit, 3-bit, and 4-bit adders. It is also demonstrated for arithmetic circuits that the error metric based on MAE provides better results than Hamming distance. This work is extended for 2-bit, 3-bit, and 4-bit multipliers and median filters in~\cite{vasicek:sekanina:tec}. Evolved small approximate multipliers are deployed as building blocks of larger 8-bit and 16-bit approximate multipliers that show better trade-offs than the approximate multipliers presented in seminal paper~\cite{Jie:R25UDM}.

CGP evolved a library of approximate arithmetic circuits that exhibit many different trade-offs between the error (expressed by several error metrics) and circuit properties. The first version of the library, EvoApprox8b, contains hundreds of approximate implementations of 8-bit adders and multipliers~\cite{Mrazek:date17:evoapprox8b}. 
To seed the initial population, the authors utilize 13 different adders and 6 different multipliers ranging from the basic implementations such as Ripple-Carry Adder or Ripple-Carry Array Multiplier to advanced architectures such as Higher Valency Tree Adder and Wallace Tree Multiplier.
Power consumption is estimated using the switching activity analysis. The delay of a candidate circuit is calculated as the sum of the delays on the cells along the longest path. The delay of a cell is modeled as a function of its input transition time and capacitive load on the cell's output. The library was later extended to EvoApproxLib and EvoApproxLib-LITE~\cite{Mrazek:JETCAS:20}. 

\begin{figure}[ht]
    \centering
    \includegraphics[width=0.49\textwidth]{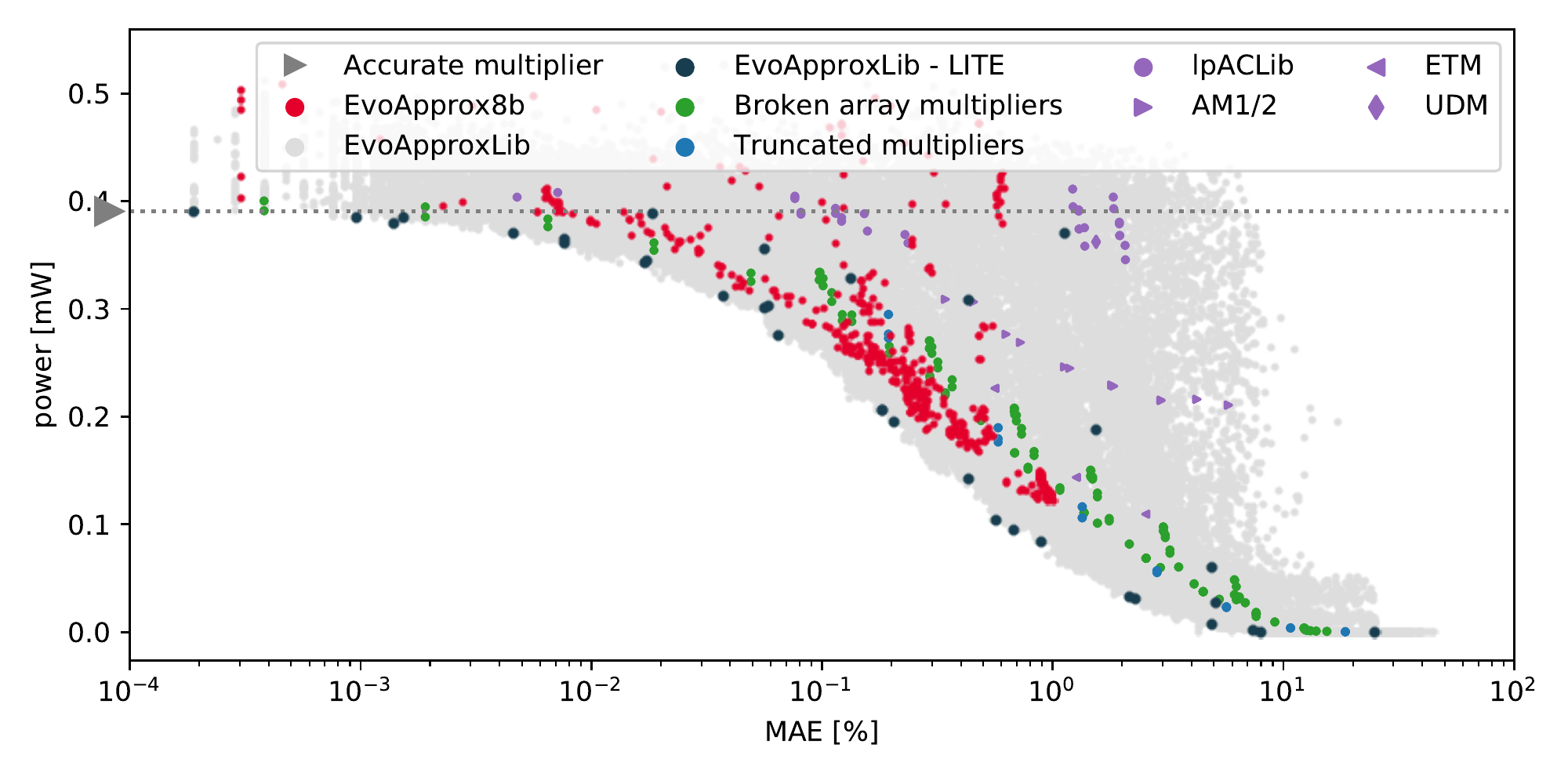}
    \caption{The MAE-power plot for 8-bit approximate multipliers taken from the EvoApproxLib-LITE (black points), EvoApproxLib (grey points), and EvoApprox8b (red points) are compared with broken array multipliers (green points), truncated multipliers (blue points) and other selected approximate multipliers from lpAClib~\cite{Shafique:dac16} and~\cite{JiangLL019:bookch}. Adopted from~\cite{Mrazek:JETCAS:20}.
    }
\label{fig:cmp}
\end{figure}

In Fig.~\ref{fig:cmp}, evolved approximate 8-bit multipliers are compared with other approximate multipliers. Because of limited space, we can show only the MAE vs. power trade-offs. The black points represent the EvoApproxLib-LITE version of the library. The original circuits of EvoApprox8b are red points, conventional broken array multipliers are green points, and truncated multipliers are blue points. Purple color denotes other multipliers analyzed in~\cite{JiangLL019:bookch} (approximate multipliers (AM), error-tolerant multipliers (ETM), and underdesigned multipliers (UDM)) and lpAClib multipliers~\cite{Shafique:dac16}. The grey points in Fig.~\ref{fig:cmp} show all 16,833 non-dominated implementations currently available in the EvoApproxLib.
All approximate multipliers were synthesized with Synopsys Design Compiler, 45 nm process, and $V_{dd}=$1V.

The EvoApproxLib was later extended with evolved approximate implementations of up to 128-bit adders, up to 32-bit multipliers, dividers, Multiply\&Accumulate (MAC) circuits, and $8\times{k}$-bit multipliers, $k \in \{4, 5,6,7\}$~\cite{Mrazek:JETCAS:20,Ceska:iccad17,Ceska:ASC:2020}. Each approximate circuit is fully characterized in terms of several error metrics, electrical parameters for various technology libraries and can be downloaded in Verilog, C, and Matlab formats.



The CGP-based approximation can provide quality-configurable circuits whose accuracy (and thus power consumption) can be dynamically switched according to the application requirements~\cite{Mrazek:AHS18}. In another research, approximate multipliers are evolved in such a way that they show low error for the most frequent input vectors. In contrast, a higher error is acceptable for the remaining input vectors~\cite{Vasicek:date2019}. This feature is essential in applications such as CNN accelerators, where the distribution of weights is known, and multipliers can be evolved with respect to the particular set of weights.  De Souza et al. compare two mutation operators in a multi-objective approximation on 15 benchmark circuits with up to 14 inputs~\cite{Souza:MOCGPApprox:2020}. Mrazek at al.~\cite{Mrazek:iet:18} investigate if the standard CGP utilizing two-input gates can be improved by supporting more complex building blocks such as full adders. They report over 10$\times$ speedup in the approximation of 12-bit multipliers if CGP can use full adders as building blocks.

Another innovative application of approximate computing is in reducing the area overhead associated with unavoidable hardware redundancy needed to ensure the reliability of digital circuits. Sanchez-Clemente et al.~\cite{TREL:16} compare two approaches to reduce the area of the Triple Modular Redundancy scheme (TMR).  The evolutionary approach based on CGP provides radically different solutions that are hard to reach by other methods.  The goal of CGP is to evolve suitable under- and over-approximations (measured using the Hamming distance) of the original circuit while the area is minimized. 
The (second) probabilistic approach approximates a circuit in a greedy manner, and the method is based on a probabilistic estimation of the error. Experimental results obtained for several combinational circuits demonstrated that the evolutionary approach can produce better solutions, but the probabilistic approach is close.
In the follow-up work, Albandes et al.~\cite{Albandes:2018} propose another approach. GA has to find replacements for some gates according to a predefined library that specifies what transformations are valid for each logic gate. For example, a two-input NAND(x,y) is underapproximated either by NOT(x) or NOT(y) or the zero constant. The candidate replacements are stored in the genotype. NSGA-II tries to find the best trade-offs between the size of the circuit (to be minimized) and the Fault Coverage (to be maximized).

\subsection{Approximate Circuits for FPGAs} 

Research on approximate logic circuits~\cite{Vasicek:fpl:16}, as well as approximate arithmetic circuits and image filters~\cite{Vas:Mra:Sek:ICES16}, revealed that approximations discovered by CGP at the gate level are almost perfectly preserved by common synthesis tools for FPGAs if these circuits are not small. On the other hand, common synthesis tools provide (for some circuits) results far from optimum. For example, a 40\% reduction (68 LUTs) was achieved by CGP for ‘clmb‘ benchmark circuit (Bus Interface) without introducing any error. Additional
43\% reduction can be obtained by introducing only a 0.1\% error~\cite{Vasicek:fpl:16}.
Recently, Vasicek~\cite{Vasicek:DDECS:21} presented a novel iterative method for the synthesis of approximate circuits optimized for LUT-based technologies. The method operates internally at the level of 2-input LUTs and produces LUT-level netlists that may consist of up to $k$-input LUTs, where $k$ can be chosen by the user depending on the intended target technology. One of the critical properties of the proposed method is the ability to keep the propagation delay under control. For various approximate adders (up to 64-bit) and multipliers (8-bit and 16-bit) the method produced better trade-offs than the state of the art when the WCE, the number of LUTs, and propagation delay are considered.

\subsection{Formal Error Analysis in the Fitness Function}
\label{sec:form:verify}

As the circuit simulation across all possible input vectors is intractable for complex circuits and with the aim of providing the exact error of an approximate implementation, two main approaches, based on Reduced Ordered Binary Decision Diagrams (ROBDD) and satisfiability (SAT) solvers, have been developed for relaxed equivalence checking~\cite{vasicek:access2019}. In both cases, an auxiliary circuit, the so-called \emph{miter}, is constructed and then analyzed. The miter connects corresponding outputs of the candidate circuit (to be checked), the golden circuit, and an error-specific circuit to determine the approximation error. 

If the auxiliary circuit is expressed using an ROBDD the requested error metrics are obtained by analyzing the ROBDD. As ROBDDs are inefficient in representing classes of circuits for which the number of nodes in ROBDD is growing exponentially with the number of input variables (e.g., multipliers and dividers), their use in equivalence checking of approximate circuits is typically possible for adders and other less structurally complex functions~\cite{vasicek:access2019}. On the other hand, all commonly used error metrics can be evaluated by means of ROBDDs. For example, the average Hamming distance is employed for approximation of complex combinational circuits at the gate-level in~\cite{vasicek:sekanina:genp16} and for FPGAs in~\cite{Vasicek:fpl:16}. The MAE is (exactly) calculated using ROBDDs for approximate adders. These adders are employed in approximate DCT circuits, which are evaluated in a hardware implementation of the HEVC video standard~\cite{Vasicek:date:hevc}.

Suppose now that the error analysis is based on SAT solving. In that case, the miter is represented as a logic formula in Conjunctive Normal Form (CNF) for which SAT solver decides whether it is satisfiable or unsatisfiable. The interpretation of this outcome depends on the construction of the miter. Common SAT solvers are, in principle, applicable to the worst-case analysis only. However, this approach is more scalable than ROBDDs and, hence, it is applied to determine WCE of complex approximate multipliers~\cite{Ceska:iccad17}, dividers, and MACs~\cite{Ceska:ASC:2020}. Here, CGP is combined with SAT solving uniquely in the so-called \emph{verifiability driven search strategy}~\cite{Ceska:iccad17}. CGP generates such candidate approximate circuits that can be evaluated by a SAT solver (i.e., the SAT solver decides whether the circuit satisfies a given error bound) within a given time limit. If it can be proved by the SAT solver that the candidate circuit satisfies the specification, its fitness score corresponds to its area, and the goal is to minimize the area. Otherwise, the circuit is not utilized anymore. 

The principles of approximate computing can be applied directly at the level of ROBDD circuit representation~\cite{ BDDapprox:gecco:17}, i.e., before any circuit implementation is carried out. ROBDDs are minimized under several objectives by performing both variable reordering and approximation while a predefined error constraint is not violated. A multi-objective GA uses an $\epsilon$-preferred relation to compare subvectors of objective functions with equal priorities. The genotype contains two parts: permutation of input variables of the given Boolean function and a vector consisting of pairs designating the approximation operators and the ROBDD level indices where each operator should be applied. On 20 Boolean functions from the ISCAS89 benchmark set, the method reduces the ROBDD size by 68.02\% on average in comparison with the initial ROBDDs. This improvement is achieved at a low cost of total inaccuracy (the average is 2.12\%), which is insignificant compared to the amount of size reduction.

\subsection{Higher Levels of Abstraction}

Instead of one-bit connections and elementary gates, CGP can work at the level of multi-bit connections and elementary functions such as adders, multipliers, and comparators. 

Approximate multiple-constant multiplierless multipliers are evolved by a multi-objective CGP~\cite{Petrlik:ddecs13}. The specification includes a set of building blocks (in particular, adders, subtractors, and shifts) and a set of $k$ constants $\{C_1, \dots, C_k\}$ for which the circuit has to output $k$ products, i.e., $C_1x, \dots, C_kx$. The goal is to minimize the average error across all the products, the number of additions/subtractions, and delay. 

By using simple functions such as addition, minimum and maximum, CGP provides approximate implementations of arctan and square root functions~\cite{Wiglasz:ICES18}. These evolved functions then replace their standard implementations in the gradient orientation and magnitude computation modules that are critical components of the gradient histogram computation of the histogram of oriented gradients (HOG) feature extraction method. This method, followed by the standard support vector machine (SVM)-based classifier, was used to detect pedestrians in images. 

Kemcha et al.~\cite{Kemcha:Nedjah:19} introduce a multi-objective Particle Swarm Optimization to evolve approximate implementations of an exact sequential circuit (in particular, a sequential divider) modeled at the RT level. Candidate circuits are represented using a vector of integers that can be interpreted as a netlist defining components and their connections. 
Contrasted to the standard CGP, feedback connections are supported in the circuit. The objective is to find the best trade-offs between accuracy, delay, and area. 

Another application is searching for approximate implementations of the median function. In hardware, the median circuit typically consists of a sequence of compare-and-swap operators over the primary inputs and intermediate results. By reducing the number of compare-and-swap operations, one can improve the energy-efficiency of median function computing, which is helpful in many applications, including image filtering and sensor data processing. The approximate implementations of 9-input and 25-input median obtained by CGP are evaluated on several microcontrollers, where the actual power consumption is measured ~\cite{Mrazek:gecco:GI:2015}. This work is extended in~\cite{Vasicek:Mrazek:genp17}, where the authors propose a new combinatorial analysis-based approach and the so-called distance error enabling to precisely determine the error for a candidate approximate median network without running the full simulation. 


\subsection{Source Code Level}

The automated approximation method called ABACUS is developed for circuits described at the behavioral or RT level in Verilog~\cite{ABACUS:DATE14,Nepal:19}. First, the source code is parsed to create its AST. Various mutation operators are then applied to the AST to derive approximate versions, which are then written back to Verilog and simulated for error evaluation and synthesized for the area and power evaluation. The mutation operators include the bit width simplifications, variable to constant substitutions, approximate arithmetic transformations (e.g., an exact addition is replaced by an approximate addition), expression transformations (e.g., \verb,z = a*b+c*d, is replaced by \verb,z = a*(b+d),), and loop transformations. While the original version of ABACUS uses an iterative stochastic greedy algorithm~\cite{ABACUS:DATE14}, the latest version is based on NSGA-II~\cite{Nepal:19}.

\subsection{Approximation From Scratch}

Rather than creating an approximate implementation of an existing solution, evolutionary algorithms sometimes enable to evolve an entirely new approximation from scratch. This approach is useful when we do not have any suitable initial solution, or the problem is relatively simple, and EA is thus directly capable of providing very competitive results. 

Towards this end, various approximate implementations of image filters (median filter, adaptive median filter, weighted median filter) are approximated by CGP, which is initialized with an exact implementation, and compared against image filters evolved from scratch~\cite{Sekanina:radioeng17}. Their quality is measured on a set of 30 images containing salt and pepper noise and impulse noise of different intensities. Evolved filters consistently outperform approximate filters, i.e., they occupy the Pareto front containing the PSNR-power trade-offs.

For network flow hashing, specialized hash functions are evolved using LGP~\cite{Grochol:ahs18}. Their quality is measured on real network data, and their implementation cost and delay are reported for an FPGA implementation. Compared to several standard hash functions, evolved hash functions provide quite competitive results.

In~\cite{GPforFEx:TC18}, GP creates programs for error-aware extraction of features from raw data. These features are then used by an SVM-based classifier in applications such as electroencephalogram-based seizure detection and electrocardiogram-based arrhythmia detection. The system is implemented on a heterogeneous processor. By controlling the size of these GP trees, one can partly control the energy requirements of the chip. The paper demonstrates a significant reduction in feature extraction energy (more than $10\times$) at the expense of a slight degradation in system performance. The hardware-related details of the chip are summarized in paper~\cite{ExploitingAFE:HW:18}. A multi-objective problem formulation together with an accelerator design was proposed in~\cite{ReducingArea:2020}.

\section{Evolutionary Approximation in CNNs}
\label{sec:nnaprox}

The unprecedented success of machine learning methods based on deep neural networks comes with the very high computation cost needed to train these networks~\cite {applicationsSurvey}. However, even a fully trained \emph{convolutional neural network} (CNN) which is used, e.g., to classify images, requires considerable resources. For example, the inference phase of a trained CNN (such as ResNet-50) requires performing $3.9 \cdot 10^9$ multiply-and-accumulate operations to classify one single input image. A lot of effort has been invested in developing compact and so energy-efficient CNN architectures~\cite{sze:pieee17}. In addition to manual design methods, the so-called \emph{neural architecture search} significantly helped to automate the neural network design process~\cite{Elsken:nas:survey:2019, NAS:ACMSurv:2021}. One of the popular approaches is the evolutionary design of CNNs. The design objective is to provide high-quality trade-offs between the network accuracy, size, and energy. Moreover, various approximations can be introduced and explored during the evolutionary design process.

CNNs usually contain from four to tens layers of different types. When used as an image classifier, the input layer provides pixels of the input image to other layers. \emph{Convolutional layers} are capable of extracting useful features from the input data. Each convolutional layer generates, by applying one or several convolutional kernels (filters), a successively higher level of abstraction of the input data, called a \emph{feature map} which preserves essential yet unique information. \emph{Pooling layers} combine, e.g., by means of averaging, a set of input values into a small number of output values to reduce the network complexity. \emph{Fully connected} layers are composed of artificial neurons; each of them sums weighted input signals (coming from a previous layer) and produces a single output. Convolutional layers and fully connected layers are typically followed by non-linear \emph{activation functions} such as sigmoid or rectified linear units (ReLU). Modern CNNs also utilize other types of layers (such as residual connections, inception, and bottleneck blocks).  The structure of the network is defined by its \emph{hyperparameters} (e.g., the number and type of layers, the number of channels, the number and size of filters, etc.). This structure also determines the number of \emph{tunable parameters} (weights). Once the CNN architecture and hyperparameters are defined, CNN can be trained (i.e., the weights are iteratively updated) to minimize a loss function. While the training algorithm involves forward as well as backward computations along with the network, the resulting trained network, when used in an application, performs only the forward computations (the inference).  Hence, in most cases, only the inference engine is implemented and accelerated for a target application. These implementations are developed for GPUs, microcontrollers, and as specialized accelerators in ASICs and FPGAs. The architecture of CNN is thus optimized concerning resources available on the target platform and to meet the target latency.  

Approximate implementations are introduced to CNNs at the level of data type selection,  quantization, microarchitecture (e.g., pruning), arithmetic circuits, and memory subsystem. The most significant gains are obtained when adopting a cross-layer approximation approach, which involves software, architecture, and hardware at the same time, breaking thus conventional methods focused on optimizing each layer of abstraction independently~\cite{Venkataramani:IEEEProc:2020}. A recent survey of methods for evolutionary design of neural network architectures was presented in~\cite{Stanley:nature:2019}. We will focus on methods that are related to evolutionary approximations. Table~\ref{tab:class:eann} shows that the evolutionary approximation can be utilized for very different purposes during the CNN design. We briefly introduce the main techniques in the following paragraphs. 

\subsection{Evolutionary Approximation of Components of NNs}

The most obvious approach is to approximate non-linear mathematical functions that have to be implemented in hardware with reduced resources. In general, this is a well-explored area of research~\cite{Muller:pieee:2020}. Recent works utilizing EAs have addressed specific hardware constraints~\cite{
EvoPNN:19, Minarik:eurogp18}. A lot of effort is also invested to obtaining suitable approximate implementations of multipliers because the multiplication and the MACs are dominating arithmetic operations in CNNs~\cite{sze:pieee17}. For CNNs with quantized weights, the 8-bit and 8x$k$-bit approximate multipliers are evolved by CGP with the goal to minimize WCE and power consumption of resulting multipliers~\cite{Mrazek:ICCAD:16,Mrazek:JETCAS:20}. Experiments reveal that the inexact multiplication by zero leads to the accumulation of errors in CNNs and poor classification accuracy. Hence, an additional constraint was formulated for CGP: if at least one of the operands is zero, the product must be precisely zero. The CNNs utilizing evolved approximate multipliers satisfying this constraint show good error-energy trade-offs for ResNet networks on CIFAR-10 dataset~\cite{Mrazek:JETCAS:20}. These evolved multipliers are also used by ALWANN tool that provides highly-optimized implementations of CNNs for custom low-power accelerators in which the number of computing units is lower than the number of layers~\cite{Mrazek:alwann:iccad19}. ALWANN is capable of selecting a suitable approximate multiplier for each computing unit from a library of approximate multipliers in such a way that (i) one approximate multiplier serves several layers, and (ii) the overall error and energy consumption are minimized. The optimizations, including the multiplier selection problem, are solved by means of the NSGA-II algorithm.

\subsection{Hardware-Aware Neural Architecture Search}

The evolutionary design of neural networks, or neuroevolution, has recently led to the fully automated design of complex CNNs that are quite competitive in terms of accuracy and size, even for the most challenging datasets such as ImageNet~\cite{NSGANetV2:2020}. In order to represent a candidate CNN in the genotype, a well-known CNN (such as MobileNetV2 in~\cite{APQ:2020}) is usually taken as a template. The genotype then contains a set of parameters, each of them specifying possible values of the critical network’s hyperparameters (the layer type, the number of filters, the kernel sizes, etc.). In some cases, the genotype also specifies internal connections of computational elements in selected layers; in other words, it encodes a computational graph of the neural network. Concerning the target hardware accelerator and latency, the EA can thus optimize not only the hyperparameters but also computational subgraphs and various quantization and approximation schemes. According to the genotype, a candidate CNN is built. It is trained on the training data set and evaluated in terms of its accuracy, size, power and other parameters. The new generation of CNNs is created using common operators used in EAs and considering the multi-objective nature of the problem. For example, the mutation operators typically enable inserting and deleting layers, channels, or modifying the number and size of filters in~\cite{SchornEVRGA20}.

In~\cite{SchornEVRGA20}, the fixed-point quantization is applied as a post-processing step after NAS is finished. However, for example, in APQ, a suitable quantization scheme is directly evolved during NAS ~\cite{APQ:2020}. APQ thus performs a joint search for architecture, pruning, and quantization policy starting with the MobileNetV2 network. To reduce the high computation overhead associated with NAS, the accuracy is predicted using a quantization-aware predictor implemented as a 3-layer feed-forward neural network.  The input to the predictor is the encoding of the network architecture, the pruning strategy, and the quantization policy. The predictor is first trained without quantization, followed by transferring its weights to train the quantization-aware predictor, largely reducing the quantization data collection time. The latency and energy of each layer are precomputed and stored in the look-up tables. Similar techniques, including weight sharing, training data reduction, accuracy and latency estimation are widely adopted to reduce the design time because the CNN design is very computationally expensive~\cite{NAS:ACMSurv:2021}. To reduce the energy needed for multiplication in convolutional layers and maximize the overall accuracy, Pinos et al.~\cite{Pinos:eurogp:21} co-evolved CNN architecture together with a suitable approximate multiplier taken from the EvoApproxLib.

A significant improvement in the search efficiency can be obtained by suitable reduction and optimization of the search space before the NAS is conducted. For example, MCUNet, which optimizes CNNs for microcontrollers, adopts a two-stage neural architecture search approach that first optimizes the search space to fit the resource constraints and then specializes the network architecture in the optimized search space~\cite{lin2020mcunet}. A similar approach, but targeting smaller microcontrollers, is presented in $\mu$NAS~\cite{muNAS:21}. HSCoNAS~\cite{hsconas:DATE21} adopts progressive space shrinking to refine the search space towards target hardware and thus reduces the search overheads. By means of NSGA-II, NASCaps~\cite{NASCaps:Vojta:2020} jointly optimizes the network accuracy and the hardware efficiency (energy, memory, and latency) of an ASIC-based hardware accelerator developed for the so-called capsule networks. 

\begin{table}[t]
    \centering
    \caption{Evolutionary algorithms in approximate computing methods applied to hardware-aware deep neural network design. The `classifier' means a CNN-based image classifier. The `hyperp.' means that EA is used to optimize CNN's hyperparameters, together with hardware parameters ($+$HW), quantization ($+$Q), and network architecture ($+$Arch.)}
    \label{tab:class:eann}
    \resizebox{\columnwidth}{!}{
    \begin{tabular}{llllll} 
\hline
{\bf Year}	&	{\bf Cite}	&	{\bf EA}	&	{\bf Level}	&	{\bf EA's target}	&	{\bf Platform}	\\
\hline
2016	& \cite{Mrazek:ICCAD:16} &	CGP	&	gate	&	8b multiplier	&	ASIC	\\
2017	& \cite{Minarik:eurogp18} &	LGP	&	instruction	&	function approx.	&	CPU	\\
2019	& \cite{EvoPNN:19} &	GA	&	vector	&	function approx.	&	FPGA	\\
2019	& \cite{Mrazek:alwann:iccad19} &	GA	&	vector	&	multiplier selection	&	ASIC	\\
2019	& \cite{SchornEVRGA20} &	GA	&	hyperp.	&	classifier	&	ASIC	\\
2019	& \cite{Margala:2019} &	GA	&	hyperp.$+$HW	&	classifier	&	FPGA	\\
2019	& \cite{ChamNet:CVPR:2019} &	GA	&	hyperp.	&	classifier	&	multiple	\\
2020	& \cite{NASCaps:Vojta:2020} &	GA	&	hyperp.	&	classifier	&	ASIC	\\
2020	& \cite{NASCaps:Vojta:2020} &	GA	&	hyperp.	&	classifier	&	ASIC	\\
2020	& \cite{APQ:2020} &	GA	&	hyperp.$+$Q	&	classifier	&	ASIC	\\
2020	& \cite{lin2020mcunet} &	GA	&	hyperp.	&	classifier	&	MCU	\\
2020	& \cite{DeepMaker:2020} &	GA	&	hyperp.	&	classifier	&	multiple	\\
2020	& \cite{Mrazek:JETCAS:20} &	CGP	&	gate	&	8x$k$-bit multiplier	&	ASIC	\\
2021	& \cite{hsconas:DATE21} &	GA	&	hyperp.	&	classifier	&	multiple	\\
2021	& \cite{muNAS:21} &	GA	&	hyperp.	&	classifier	&	MCU	\\
2021	& \cite{Pinos:eurogp:21} &	CGP	&	hyperp.$+$Arch &	classifier	&	ASIC	\\

\hline
\end{tabular}
    }
\end{table}

\section{Conclusions}
\label{sec:conclusions}
We provided the first survey of EA-based approaches used in approximate computing. From over 60 papers dealing with this topic, we learned that EAs are applied as multi-objective optimizers, with the aim not only to optimize some parameters of approximate HW/SW systems but also to determine their architecture. EAs were used for all relevant design abstractions and in tens of different applications. However, as EAs suffer from long execution times, their utilization has (almost always) to be accompanied by a suitable fitness estimation strategy. This observation is especially valid for a newly established research topic – the hardware-aware neural architecture search helping in the automated design of approximate accelerators of deep neural networks.

\setcounter{secnumdepth}{0}
\section{Acknowledgements}
This work was supported by the Czech science foundation project 21-13001S.

\bibliographystyle{IEEEtran}
\bibliography{IEEEabrv,0_ref}

\begin{thebibliography}{10}
\providecommand{\url}[1]{#1}
\csname url@samestyle\endcsname
\providecommand{\newblock}{\relax}
\providecommand{\bibinfo}[2]{#2}
\providecommand{\BIBentrySTDinterwordspacing}{\spaceskip=0pt\relax}
\providecommand{\BIBentryALTinterwordstretchfactor}{4}
\providecommand{\BIBentryALTinterwordspacing}{\spaceskip=\fontdimen2\font plus
\BIBentryALTinterwordstretchfactor\fontdimen3\font minus
  \fontdimen4\font\relax}
\providecommand{\BIBforeignlanguage}[2]{{%
\expandafter\ifx\csname l@#1\endcsname\relax
\typeout{** WARNING: IEEEtran.bst: No hyphenation pattern has been}%
\typeout{** loaded for the language `#1'. Using the pattern for}%
\typeout{** the default language instead.}%
\else
\language=\csname l@#1\endcsname
\fi
#2}}
\providecommand{\BIBdecl}{\relax}
\BIBdecl

\bibitem{Hipeac:visio:2021}
\BIBentryALTinterwordspacing
M.~Duranton, K.~D. Bosschere, B.~Coppens, C.~Gamrat, T.~Hoberg, H.~Munk,
  C.~Roderick, T.~Vardanega, and O.~Zendra, \emph{{HiPEAC} Vision 2021}, 2021.
  [Online]. Available: \url{https://www.hipeac.net/vision}
\BIBentrySTDinterwordspacing

\bibitem{Mittal:2016}
S.~Mittal, ``A survey of techniques for approximate computing,'' \emph{ACM
  Comput. Surv.}, vol.~48, no.~4, p. 1–33, 2016.

\bibitem{ACsurvey:ACM:2020}
P.~Stanley-Marbell, A.~Alaghi, M.~Carbin, E.~Darulova, L.~Dolecek,
  A.~Gerstlauer, G.~Gillani, D.~Jevdjic, T.~Moreau, M.~Cacciotti, A.~Daglis,
  N.~E. Jerger, B.~Falsafi, S.~Misailovic, A.~Sampson, and D.~Zufferey,
  ``Exploiting errors for efficiency: {A} survey from circuits to
  applications,'' \emph{ACM Comput. Surv.}, vol.~53, no.~3, 2020.

\bibitem{Scarabottolo:pieee:2020}
I.~Scarabottolo, G.~Ansaloni, G.~A. Constantinides, L.~Pozzi, and S.~Reda,
  ``Approximate logic synthesis: {A} survey,'' \emph{Proceedings of the IEEE},
  vol. 108, no.~12, pp. 2195--2213, 2020.

\bibitem{Sekanina:bookch:19}
L.~Sekanina, Z.~Vasicek, and V.~Mrazek, ``Automated search-based functional
  approximation for digital circuits,'' in \emph{Approximate Circuits,
  Methodologies and {CAD}}, S.~Reda and M.~Shafique, Eds.\hskip 1em plus 0.5em
  minus 0.4em\relax Springer, 2019, pp. 175--203.

\bibitem{Hadi:CACM:2015}
H.~Esmaeilzadeh, A.~Sampson, L.~Ceze, and D.~Burger, ``Neural acceleration for
  general-purpose approximate programs,'' \emph{Commun. ACM}, vol.~58, no.~1,
  p. 105–115, 2014.

\bibitem{Mazahir:2017aa}
S.~Mazahir, O.~Hasan, R.~Hafiz, M.~Shafique, and J.~Henkel, ``Probabilistic
  error modeling for approximate adders,'' \emph{IEEE Transactions on
  Computers}, vol.~66, no.~3, pp. 515--530, 2017.

\bibitem{vasicek:access2019}
Z.~Vasicek, ``Formal methods for exact analysis of approximate circuits,''
  \emph{IEEE Access}, vol.~7, no.~1, pp. 177\,309--177\,331, 2019.

\bibitem{EibenS15}
A.~E. Eiben and J.~E. Smith, \emph{Introduction to Evolutionary Computing,
  Second Edition}, ser. Natural Computing Series.\hskip 1em plus 0.5em minus
  0.4em\relax Springer, 2015.

\bibitem{poli08:fieldguide}
R.~Poli, W.~B. Langdon, and N.~F. McPhee, \emph{A field guide to genetic
  programming}.\hskip 1em plus 0.5em minus 0.4em\relax lulu.com, 2008.

\bibitem{Miller:cgp:2020}
J.~F. Miller, ``Cartesian genetic programming: its status and future,''
  \emph{Genet. Program. Evolvable Mach.}, vol.~21, no. 1-2, pp. 129--168, 2020.

\bibitem{moea:2020}
C.~A. Coello~Coello, S.~Gonzalez~Brambila, J.~Figueroa~Gamboa, M.~G.
  Castillo~Tapia, and R.~Hernandez~Gomez, ``Evolutionary multiobjective
  optimization: open research areas and some challenges lying ahead,''
  \emph{Complex \& Intelligent Systems}, vol. 2020, pp. 1--16, 2020.

\bibitem{deb2002}
K.~Deb, A.~Pratap, S.~Agarwal, and T.~Meyarivan, ``A fast and elitist
  multiobjective genetic algorithm: {NSGA-II},'' \emph{IEEE Transactions on
  Evolutionary Computation}, vol.~6, no.~2, pp. 182--197, 2002.

\bibitem{Ansel:2011}
J.~Ansel, Y.~L. Wong, C.~Chan, M.~Olszewski, A.~Edelman, and S.~Amarasinghe,
  ``Language and compiler support for auto-tuning variable-accuracy
  algorithms,'' in \emph{International Symposium on Code Generation and
  Optimization (CGO 2011)}, 2011, pp. 85--96.

\bibitem{Park2014ExpAXAF}
\BIBentryALTinterwordspacing
J.~Park, X.~Zhang, K.~Ni, H.~Esmaeilzadeh, and M.~Naik, ``{ExpAX: A} framework
  for automating approximate programming,'' 2014. [Online]. Available:
  \url{https://smartech.gatech.edu/handle/1853/52032}
\BIBentrySTDinterwordspacing

\bibitem{demara:15}
A.~A. Naseer, R.~A. Ashraf, D.~Dechev, and R.~F. DeMara, ``Designing
  energy-efficient approximate adders using parallel genetic algorithms,'' in
  \emph{SoutheastCon 2015}.\hskip 1em plus 0.5em minus 0.4em\relax IEEE, 2015,
  pp. 1--7.

\bibitem{Grater:DATE16}
A.~Lotfi, A.~Rahimi, A.~Yazdanbakhsh, H.~Esmaeilzadeh, and R.~K. Gupta,
  ``Grater: An approximation workflow for exploiting data-level parallelism in
  fpga acceleration,'' in \emph{2016 Design, Automation and Test in Europe,
  Conference and Exhibition (DATE)}, 2016, pp. 1279--1284.

\bibitem{Vaverka:CompLib:16}
F.~Vaverka, R.~Hrbacek, and L.~Sekanina, ``Evolving component library for
  approximate high level synthesis,'' in \emph{2016 IEEE Symposium Series on
  Computational Intelligence (SSCI)}, 2016, pp. 1--8.

\bibitem{ApproxCompressSens:2017}
S.~P. Kadiyala, V.~K. Pudi, and S.-K. Lam, ``Approximate compressed sensing for
  hardware-efficient image compression,'' in \emph{30th IEEE International
  System-on-Chip Conference}, 2017, pp. 340--345.

\bibitem{ApproxMultNair:18}
I.~Alouani, H.~Ahangari, O.~Ozturk, and S.~Niar, ``A novel heterogeneous
  approximate multiplier for low power and high performance,'' \emph{IEEE
  Embedded Systems Letters}, vol.~10, no.~2, pp. 45--48, 2018.

\bibitem{Albandes:2018}
I.~Albandes, A.~Serrano-Cases, A.~Sanchez-Clemente, M.~Martins,
  A.~Martinez-Alvarez, S.~Cuenca-Asensi, and F.~L. Kastensmidt, ``Improving
  approximate-{TMR} using multi-objective optimization genetic algorithm,'' in
  \emph{2018 IEEE 19th Latin-American Test Symposium (LATS)}, 2018, pp. 1--6.

\bibitem{Kadiyala:19}
S.~P. Kadiyala, A.~Sen, S.~Mahajan, Q.~Wang, A.~Lingamaneni, J.~S. German,
  X.~Hong, K.~V. Palem, and A.~Basu, ``An optimum inexact design for an energy
  efficient hearing aid,'' \emph{J. Low Power Electron.}, vol.~15, no.~2, pp.
  129--143, 2019.

\bibitem{CANDARW:20}
N.~A. Vu~Doan, M.~Manuel, S.~Conrady, A.~Kreddig, and W.~Stechele, ``Parameter
  optimization of approximate image processing algorithms in {FPGAs},'' in
  \emph{2020 Eighth International Symposium on Computing and Networking
  Workshops (CANDARW)}, 2020, pp. 74--80.

\bibitem{WorkloadAware:DATE2021}
D.~Ma, R.~Thapa, X.~Wang, C.~Hao, and X.~Jiao, ``Workload-aware approximate
  computing configuration,'' in \emph{Design Automation and Test in
  Europe}.\hskip 1em plus 0.5em minus 0.4em\relax EDAA, 2021, pp. 920--923.

\bibitem{BarbareschiBM21}
\BIBentryALTinterwordspacing
M.~Barbareschi, S.~Barone, and N.~Mazzocca, ``Advancing synthesis of decision
  tree-based multiple classifier systems: an approximate computing case
  study,'' \emph{Knowl. Inf. Syst.}, vol.~63, no.~6, pp. 1577--1596, 2021.
  [Online]. Available: \url{https://doi.org/10.1007/s10115-021-01565-5}
\BIBentrySTDinterwordspacing

\bibitem{Barone:Access:21}
S.~Barone, M.~Traiola, M.~Barbareschi, and A.~Bosio, ``Multi-objective
  application-driven approximate design method,'' \emph{IEEE Access}, vol.~9,
  pp. 86\,975--86\,993, 2021.

\bibitem{Petrlik:ddecs13}
J.~Petrlik and L.~Sekanina, ``Multiobjective evolution of approximate multiple
  constant multipliers,'' in \emph{2013 IEEE 16th International Symposium on
  Design and Diagnostics of Electronic Circuits Systems (DDECS)}, 2013, pp.
  116--119.

\bibitem{sekanina:axc:ices13}
L.~Sekanina and Z.~Vasicek, ``Approximate circuit design by means of evolvable
  hardware,'' in \emph{2013 IEEE International Conference on Evolvable Systems
  (ICES)}, 2013, pp. 21--28.

\bibitem{vasicek:sekanina:tec}
Z.~Vasicek and L.~Sekanina, ``Evolutionary approach to approximate digital
  circuits design,'' \emph{IEEE Transactions on Evolutionary Computation},
  vol.~19, no.~3, pp. 432--444, 2015.

\bibitem{Mrazek:gecco:GI:2015}
V.~Mrazek, Z.~Vasicek, and L.~Sekanina, ``Evolutionary approximation of
  software for embedded systems: {M}edian function,'' in \emph{Proceedings of
  the Companion Publication of the 2015 Annual Conference on Genetic and
  Evolutionary Computation}, ser. GECCO Companion '15.\hskip 1em plus 0.5em
  minus 0.4em\relax New York, NY, USA: ACM, 2015, pp. 795--801.

\bibitem{vasicek:sekanina:genp16}
Z.~Vasicek and L.~Sekanina, ``Evolutionary design of complex approximate
  combinational circuits,'' \emph{Genetic Programming and Evolvable Machines},
  vol.~17, no.~2, pp. 1--24, 2016.

\bibitem{Vas:Mra:Sek:ICES16}
Z.~Vasicek, V.~Mrazek, and L.~Sekanina, ``Evolutionary functional approximation
  of circuits implemented into {FPGAs},'' in \emph{2016 IEEE Symposium Series
  on Computational Intelligence (SSCI)}, 2016, pp. 1--8.

\bibitem{TREL:16}
A.~J. Sanchez-Clemente, L.~Entrena, R.~Hrbacek, and L.~Sekanina, ``Error
  mitigation using approximate logic circuits: {A} comparison of probabilistic
  and evolutionary approaches,'' \emph{IEEE Transactions on Reliability},
  vol.~65, no.~4, pp. 1871--1883, 2016.

\bibitem{Vasicek:fpl:16}
Z.~Vasicek and L.~Sekanina, ``Search-based synthesis of approximate circuits
  implemented into {FPGAs},'' in \emph{26th Int. Conference on Field
  Programmable Logic and Applications}, 2016, pp. 1--4.

\bibitem{BDDapprox:gecco:17}
S.~Shirinzadeh, M.~Soeken, D.~Gro\ss{}e, and R.~Drechsler, ``An adaptive
  prioritized $\epsilon$-preferred evolutionary algorithm for approximate {BDD}
  optimization,'' in \emph{Proc. of the Genetic and Evolutionary Computation
  Conference}, ser. GECCO'17.\hskip 1em plus 0.5em minus 0.4em\relax ACM, 2017,
  pp. 1232--1239.

\bibitem{axcdct:17}
A.~Azaraien, B.~Djalaei, and M.~E. Salehi, ``Evolutionary architecture design
  for approximate {DCT},'' in \emph{19th International Symposium on Computer
  Architecture and Digital Systems}, 2017, pp. 1--2.

\bibitem{Vasicek:Mrazek:genp17}
Z.~Vasicek and V.~Mrazek, ``Trading between quality and non-functional
  properties of median filter in embedded systems,'' \emph{Genetic Prog. and
  Evolvable Machines}, vol.~18, no.~1, pp. 45--82, 2017.

\bibitem{Vasicek:date:hevc}
Z.~Vasicek, V.~Mrazek, and L.~Sekanina, ``Towards low power approximate {DCT}
  architecture for {HEVC} standard,'' in \emph{Design, Automation {\&} Test in
  Europe Conference {\&} Exhibition, {DATE} 2017}, 2017, pp. 1576--1581.

\bibitem{Mrazek:date17:evoapprox8b}
V.~{Mrazek}, R.~{Hrbacek}, Z.~{Vasicek}, and L.~{Sekanina}, ``Evoapprox8b:
  Library of approximate adders and multipliers for circuit design and
  benchmarking of approximation methods,'' in \emph{Design, Automation Test in
  Europe Conference Exhibition, 2017}, 2017, pp. 258--261.

\bibitem{Ceska:iccad17}
M.~Ceska, J.~Matyas, V.~Mrazek, L.~Sekanina, Z.~Vasicek, and T.~Vojnar,
  ``Approximating complex arithmetic circuits with formal error guarantees:
  32-bit multipliers accomplished,'' in \emph{IEEE/ACM International Conference
  on Computer-Aided Design}, 2017, pp. 416--423.

\bibitem{Sekanina:radioeng17}
L.~Sekanina, Z.~Vasicek, and V.~Mrazek, ``Approximate circuits in low-power
  image and video processing: {T}he approximate median filter,''
  \emph{Radioengineering}, vol.~26, no.~3, pp. 623--632, 2017.

\bibitem{Grochol:ahs18}
D.~Grochol and L.~Sekanina, ``Fast reconfigurable hash functions for network
  flow hashing in {FPGAs},'' in \emph{Proceedings of the 2018 NASA/ESA
  Conference on Adaptive Hardware and Systems}.\hskip 1em plus 0.5em minus
  0.4em\relax IEEE, 2018, pp. 257--263.

\bibitem{GPforFEx:TC18}
J.~Lu, H.~Jia, N.~Verma, and N.~K. Jha, ``Genetic programming for
  energy-efficient and energy-scalable approximate feature computation in
  embedded inference systems,'' \emph{IEEE Transactions on Computers}, vol.~67,
  no.~2, pp. 222--236, 2018.

\bibitem{Wiglasz:ICES18}
M.~Wiglasz and L.~Sekanina, ``Cooperative coevolutionary approximation in
  {HOG}-based human detection embedded system,'' in \emph{IEEE Symp. Series on
  Computational Intelligence}, 2018, pp. 1313--1320.

\bibitem{Mrazek:iet:18}
V.~Mrazek, Z.~Vasicek, and R.~Hrbacek, ``Role of circuit representation in
  evolutionary design of energy-efficient approximate circuits,'' \emph{IET
  Computers \& Digital Techniques}, vol. 2018, no.~4, pp. 139--149, 2018.

\bibitem{Mrazek:AHS18}
V.~Mrazek, Z.~Vasicek, and L.~Sekanina, ``Design of quality-configurable
  approximate multipliers suitable for dynamic environment,'' in \emph{2018
  NASA/ESA Conference on Adaptive Hardware and Systems (AHS)}, 2018, pp.
  264--271.

\bibitem{Kemcha:Nedjah:19}
R.~Kemcha, N.~Nedjah, A.~R. Maouche, and M.~Bougherara, ``Evolutionary design
  of approximate sequential circuits at {RTL} using particle swarm
  optimization,'' in \emph{Computational Science and Its Applications -- ICCSA
  2019}.\hskip 1em plus 0.5em minus 0.4em\relax Cham: Springer, 2019, pp.
  671--684.

\bibitem{Vasicek:date2019}
Z.~Vasicek, V.~Mrazek, and L.~Sekanina, ``Automated circuit approximation
  method driven by data distribution,'' in \emph{Design, Automation and Test in
  Europe Conference}.\hskip 1em plus 0.5em minus 0.4em\relax EDAA, 2019, pp.
  96--101.

\bibitem{ReducingArea:2020}
Y.~Tang, H.~Jia, and N.~Verma, ``Reducing energy of approximate feature
  extraction in heterogeneous architectures for sensor inference via
  energy-aware genetic programming,'' \emph{IEEE Transactions on Circuits and
  Systems I: Regular Papers}, vol.~67, no.~5, pp. 1576--1587, 2020.

\bibitem{Souza:MOCGPApprox:2020}
L.~Souza and H.~Bernardino, ``On the analysis of mutation operators in
  multiobjective cartesian genetic programming for designing combinational
  logic circuits,'' in \emph{Anais do XVII Encontro Nacional de Inteligencia
  Artificial e Computacional}.\hskip 1em plus 0.5em minus 0.4em\relax Porto
  Alegre, RS, Brasil: SBC, 2020, pp. 390--401.

\bibitem{Ceska:ASC:2020}
M.~Ceska, J.~Matyas, V.~Mrazek, L.~Sekanina, Z.~Vasicek, and T.~Vojnar,
  ``Adaptive verifiability-driven strategy for evolutionary approximation of
  arithmetic circuits,'' \emph{Applied Soft Computing}, vol.~95, no. 106466,
  pp. 1--17, 2020.

\bibitem{Vasicek:DDECS:21}
Z.~Vasicek, ``Synthesis of approximate circuits for {LUT}-based {FPGAs},'' in
  \emph{2021 24th International Symposium on Design and Diagnostics of
  Electronic Circuits Systems (DDECS)}, 2021, pp. 17--22.

\bibitem{EnerJ:SampsonDFGCG11}
A.~Sampson, W.~Dietl, E.~Fortuna, D.~Gnanapragasam, L.~Ceze, and D.~Grossman,
  ``{EnerJ}: approximate data types for safe and general low-power
  computation,'' in \emph{Proc. of the 32nd {ACM} {SIGPLAN} Conference on
  Programming Language Design and Implementation}, M.~W. Hall and D.~A. Padua,
  Eds.\hskip 1em plus 0.5em minus 0.4em\relax {ACM}, 2011, pp. 164--174.

\bibitem{Jie:R25UDM}
P.~{Kulkarni}, P.~{Gupta}, and M.~{Ercegovac}, ``Trading accuracy for power
  with an underdesigned multiplier architecture,'' in \emph{2011 24th
  Internatioal Conference on VLSI Design}, Jan 2011, pp. 346--351.

\bibitem{Mrazek:JETCAS:20}
V.~Mrazek, L.~Sekanina, and Z.~Vasicek, ``Libraries of approximate circuits:
  {A}utomated design and application in {CNN} accelerators,'' \emph{IEEE
  Journal on Emerging and Selected Topics in Circuits and Systems}, vol.~10,
  no.~4, pp. 406--418, 2020.

\bibitem{JiangLL019:bookch}
H.~Jiang, L.~Liu, F.~Lombardi, and J.~Han, ``Approximate arithmetic circuits:
  {D}esign and evaluation,'' in \emph{Approximate Circuits, Methodologies and
  {CAD}}, S.~Reda and M.~Shafique, Eds.\hskip 1em plus 0.5em minus 0.4em\relax
  Springer, 2019, pp. 67--98.

\bibitem{Shafique:dac16}
M.~Shafique, R.~Hafiz, S.~Rehman \emph{et~al.}, ``Invited: {C}ross-layer
  approximate computing: From logic to architectures,'' in \emph{DAC'16}, 2016.

\bibitem{ABACUS:DATE14}
K.~Nepal, Y.~Li, R.~I. Bahar, and S.~Reda, ``{ABACUS: A} technique for
  automated behavioral synthesis of approximate computing circuits,'' in
  \emph{Proc. of DATE'14}.\hskip 1em plus 0.5em minus 0.4em\relax EDA
  Consortium, 2014, pp. 1--6.

\bibitem{Nepal:19}
K.~Nepal, S.~Hashemi, H.~Tann, R.~I. Bahar, and S.~Reda, ``Automated high-level
  generation of low-power approximate computing circuits,'' \emph{IEEE
  Transactions on Emerging Topics in Computing}, vol.~7, no.~1, pp. 18--30,
  2019.

\bibitem{ExploitingAFE:HW:18}
H.~Jia and N.~Verma, ``Exploiting approximate feature extraction via genetic
  programming for hardware acceleration in a heterogeneous microprocessor,''
  \emph{IEEE Journal of Solid-State Circuits}, vol.~53, no.~4, pp. 1016--1027,
  2018.

\bibitem{applicationsSurvey}
S.~Dong, P.~Wang, and K.~Abbas, ``A survey on deep learning and its
  applications,'' \emph{Computer Science Review}, vol.~40, p. 100379, 2021.

\bibitem{sze:pieee17}
V.~Sze, Y.~Chen, T.~Yang, and J.~S. Emer, ``Efficient processing of deep neural
  networks: {A} tutorial and survey,'' \emph{Proceedings of the IEEE}, vol.
  105, no.~12, pp. 2295--2329, 2017.

\bibitem{Elsken:nas:survey:2019}
T.~Elsken, J.~H. Metzen, and F.~Hutter, ``Neural architecture search: {A}
  survey,'' \emph{J. Mach. Learn. Res.}, vol.~20, pp. 55:1--55:21, 2019.

\bibitem{NAS:ACMSurv:2021}
P.~Ren, Y.~Xiao, X.~Chang, P.-y. Huang, Z.~Li, X.~Chen, and X.~Wang, ``A
  comprehensive survey of neural architecture search: {C}hallenges and
  solutions,'' \emph{ACM Comput. Surv.}, vol.~54, no.~4, 2021.

\bibitem{Venkataramani:IEEEProc:2020}
S.~{Venkataramani} \emph{et~al.}, ``Efficient {AI} system design with
  cross-layer approximate computing,'' \emph{Proceedings of the IEEE}, vol.
  108, no.~12, pp. 2232--2250, 2020.

\bibitem{Stanley:nature:2019}
K.~O. Stanley, J.~Clune, J.~Lehman1, and R.~Miikkulainen, ``Designing neural
  networks through neuroevolution,'' \emph{Nature Machine Intelligence},
  vol.~1, pp. 24--35, 2019.

\bibitem{Muller:pieee:2020}
J.-M. Muller, ``Elementary functions and approximate computing,''
  \emph{Proceedings of the IEEE}, vol. 108, no.~12, pp. 2136--2149, 2020.

\bibitem{EvoPNN:19}
C.-Y. Chen, C.-W. Chang, and Z.-C. Chen, ``An evolutionary computation approach
  for approximate computing of {PNN} hardware circuits,'' in \emph{2019
  International Symposium on Intelligent Signal Processing and Communication
  Systems (ISPACS)}, 2019, pp. 1--2.

\bibitem{Minarik:eurogp18}
M.~Minarik and L.~Sekanina, ``On evolutionary approximation of sigmoid function
  for {HW/SW} embedded systems,'' in \emph{EuroGP'17}, ser. LNCS, vol.
  10196.\hskip 1em plus 0.5em minus 0.4em\relax Springer, 2017, pp. 343--358.

\bibitem{Mrazek:ICCAD:16}
V.~Mrazek, S.~S. Sarwar, L.~Sekanina, Z.~Vasicek, and K.~Roy, ``Design of
  power-efficient approximate multipliers for approximate artificial neural
  networks,'' in \emph{2016 IEEE/ACM International Conference on Computer-Aided
  Design (ICCAD)}, 2016, pp. 1--7.

\bibitem{Mrazek:alwann:iccad19}
V.~Mrazek, Z.~Vasicek, L.~Sekanina, M.~A. Hanif, and M.~Shafique, ``{ALWANN:
  A}utomatic layer-wise approximation of deep neural network accelerators
  without retraining,'' in \emph{2019 IEEE/ACM International Conference on
  Computer-Aided Design}, 2019, pp. 1--8.

\bibitem{NSGANetV2:2020}
Z.~Lu, K.~Deb, E.~Goodman, W.~Banzhaf, and V.~N. Boddeti, ``{NSGANetV2:}
  evolutionary multi-objective surrogate-assisted neural architecture search,''
  in \emph{Computer Vision -- {ECCV 2020}}.\hskip 1em plus 0.5em minus
  0.4em\relax Springer International Publishing, 2020, pp. 35--51.

\bibitem{APQ:2020}
T.~Wang, K.~Wang, H.~Cai, J.~Lin, Z.~Liu, H.~Wang, Y.~Lin, and S.~Han, ``{APQ:
  J}oint search for network architecture, pruning and quantization policy,'' in
  \emph{2020 IEEE/CVF Conference on Computer Vision and Pattern Recognition
  (CVPR)}, 2020, pp. 2075--2084.

\bibitem{SchornEVRGA20}
C.~Schorn, T.~Elsken, S.~Vogel, A.~Runge, A.~Guntoro, and G.~Ascheid,
  ``Automated design of error-resilient and hardware-efficient deep neural
  networks,'' \emph{Neural Comput. Appl.}, vol.~32, no.~24, pp.
  18\,327--18\,345, 2020.

\bibitem{Pinos:eurogp:21}
M.~Pinos, V.~Mrazek, and L.~Sekanina, ``Evolutionary neural architecture search
  supporting approximate multipliers,'' in \emph{EuroGP'21}, ser. LNCS.\hskip
  1em plus 0.5em minus 0.4em\relax Springer, 2021, pp. 82--97.

\bibitem{lin2020mcunet}
J.~Lin, W.-M. Chen, Y.~Lin, J.~Cohn, C.~Gan, and S.~Han, ``{MCUNet: T}iny deep
  learning on {IoT} devices,'' in \emph{34th Conference on Neural Information
  Processing Systems (NeurIPS 2020)}, 2020, pp. 1--12.

\bibitem{muNAS:21}
E.~Liberis, L.~Dudziak, and N.~D. Lane, ``$\mu${NAS: C}onstrained neural
  architecture search for microcontrollers,'' in \emph{Proc. of the 1st
  Workshop on Machine Learning and Systems}, ser. EuroMLSys '21.\hskip 1em plus
  0.5em minus 0.4em\relax New York, NY, USA: ACM, 2021, p. 70–79.

\bibitem{hsconas:DATE21}
X.~Luo, D.~Liu, S.~Huai, and W.~Liu, ``{HSCoNAS}: Hardware-software co-design
  of efficient dnns via neural architecture search,'' in \emph{Design,
  Automation {\&} Test in Europe Conference {\&} Exhibition, {DATE}
  2021}.\hskip 1em plus 0.5em minus 0.4em\relax {IEEE}, 2021, pp. 418--421.

\bibitem{NASCaps:Vojta:2020}
A.~Marchisio, A.~Massa, V.~Mrazek, B.~Bussolino, M.~Martina, and M.~Shafique,
  ``{NASCaps: A} framework for neural architecture search to optimize the
  accuracy and hardware efficiency of convolutional capsule networks,'' in
  \emph{2020 IEEE/ACM International Conference On Computer Aided Design
  (ICCAD)}, 2020, pp. 1--9.

\bibitem{Margala:2019}
P.~Colangelo, O.~Segal, A.~Speicher, and M.~Margala, ``Artificial neural
  network and accelerator co-design using evolutionary algorithms,'' in
  \emph{2019 IEEE High Performance Extreme Computing Conference (HPEC)}, 2019,
  pp. 1--8.

\bibitem{ChamNet:CVPR:2019}
X.~Dai, P.~Zhang, B.~Wu, H.~Yin, F.~Sun, Y.~Wang, M.~Dukhan, Y.~Hu, Y.~Wu,
  Y.~Jia, P.~Vajda, M.~Uyttendaele, and N.~K. Jha, ``{ChamNet: T}owards
  efficient network design through platform-aware model adaptation,'' in
  \emph{2019 IEEE/CVF Conference on Computer Vision and Pattern Recognition
  (CVPR)}, 2019, pp. 11\,390--11\,399.

\bibitem{DeepMaker:2020}
M.~Loni, S.~Sinaei, A.~Zoljodi, M.~Daneshtalab, and M.~Sjödin, ``{DeepMaker:
  A} multi-objective optimization framework for deep neural networks in
  embedded systems,'' \emph{Microprocessors and Microsystems}, vol.~73, p.
  102989, 2020.

\end{thebibliography}
\end{document}